\documentclass[10pt,twocolumn,letterpaper]{article}
\usepackage[accsupp]{axessibility} 
\usepackage{cvpr}
\usepackage{times}
\usepackage{epsfig}
\usepackage{graphicx}
\usepackage{amsmath}
\usepackage{amssymb}
\usepackage{graphicx}
\usepackage{amsmath}
\usepackage{amssymb}
\usepackage{booktabs}
\usepackage{tabu}

\usepackage[pagebackref]{hyperref}
\usepackage{xcolor}
\usepackage{array}
\usepackage{epsfig}
\usepackage{epstopdf}
\usepackage{booktabs}
\usepackage{algorithm}
\usepackage{algpseudocode}
\usepackage{lipsum}
\usepackage{multirow}
\usepackage{textcomp}
\usepackage{gensymb}
\usepackage{pifont}

\newcommand\blfootnote[1]{%
\begingroup 
\renewcommand\thefootnote{}\footnote{#1}%
\addtocounter{footnote}{-1}%
\endgroup 
}

\begin{document}

\title{GaitRef: Gait Recognition with Refined Sequential Skeletons}

\author{Haidong Zhu$^*$\ \ \ \ \ Wanrong Zheng$^*$\ \ \ \ \ Zhaoheng Zheng\ \ \ \ \ Ram Nevatia\\
University of Southern California\\
{\tt\small \{haidongz|wanrongz|zhaoheng.zheng|nevatia@usc.edu\}}
}

\maketitle
\thispagestyle{empty}

\begin{abstract}
Identifying humans with their walking sequences, known as gait recognition, is a useful biometric understanding task as it can be observed from a long distance and does not require cooperation from the subject. Two common modalities used for representing the walking sequence of a person are silhouettes and joint skeletons. Silhouette sequences, which record the boundary of the walking person in each frame, may suffer from the variant appearances from carried-on objects and clothes of the person. Framewise joint detections are noisy and introduce some jitters that are not consistent with sequential detections. In this paper, we combine the silhouettes and skeletons and refine the framewise joint predictions for gait recognition. With temporal information from the silhouette sequences, we show that the refined skeletons can improve gait recognition performance without extra annotations. We compare our methods on four public datasets, CASIA-B, OUMVLP, Gait3D and GREW, and show state-of-the-art performance.
\blfootnote{* Equal contribution}
\end{abstract}

\section{Introduction}

\begin{figure}[t]
    \centering
    \includegraphics[width=\linewidth]{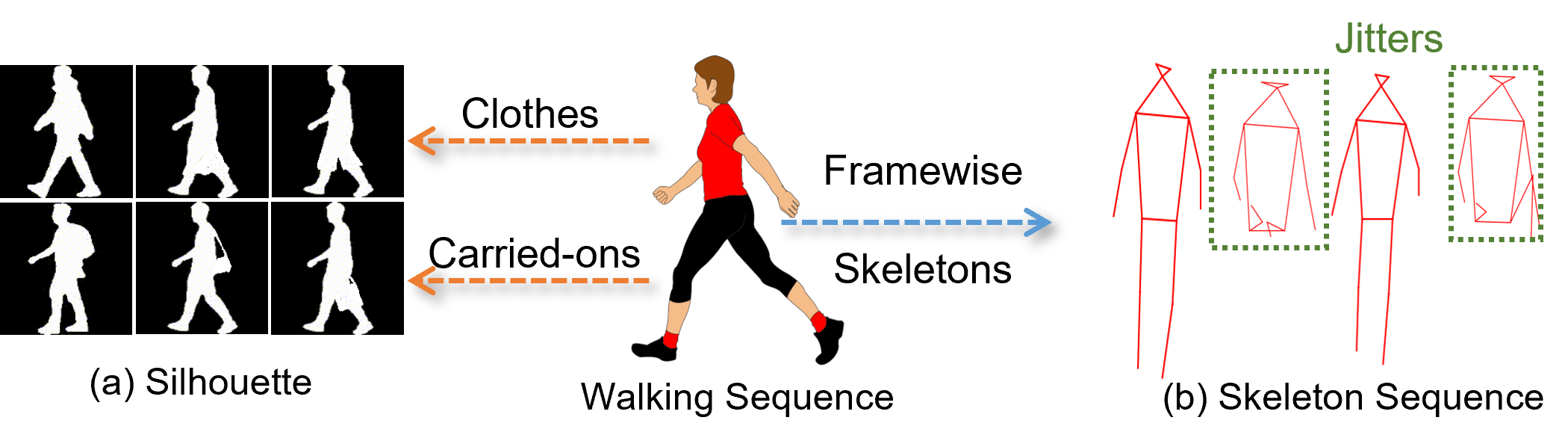}
    \caption{Visualization of the (a) silhouette and (b) skeleton sequence used for gait recognition. Silhouettes show different contours with different clothes and carried-on objects, while the skeletons suffer from jittery detection results in the video.}
    \label{fig:example}
\end{figure}

Gait recognition \cite{he2018multi,song2019gaitnet,wu2016comprehensive,yu2017invariant} aims to find the uniqueness of the walking and posture sequence of a person, which has the advantage of being able to be acquired from long distance and without the subject's cooperation. To recognize the gait sequence of a person, researchers have developed silhouettes-based methods, such as GaitSet \cite{chao2019gaitset}, GaitPart \cite{fan2020gaitpart} and GaitGL \cite{lin2021gaitgl}, and skeleton-based methods, such as GaitGraph \cite{teepe2021gaitgraph}.
However, both input modalities have some deficiencies. For binarized silhouettes, variations in clothes and carried-on objects, as shown in Figure~\ref{fig:example} (a), introduce external ambiguity, while jitters in joint detection, as Figure~\ref{fig:example} (b), decrease the skeleton accuracy.

In this paper, we introduce combination of silhouette sequences with skeletons and gaining the benefits of both modalities via refining the framewise skeletons with silhouette sequences. 
Since jitter in the detected skeletons are of a few frames isolated from the whole sequence, it is not temporally consistent with their neighbor frames \cite{zeng2021smoothnet}. Naive temporal smoothing, however, will introduce more confusion for  gait recognition since the generated skeletons create new poses not consistent with the current sequence. Meanwhile, silhouettes for neighbor frames are of better temporal consistency due to the small changes in neighbor image conditions.
We improve the quality of input skeletons by using silhouettes to fix the jitters while preserving necessary identity information for more precise gait recognition.

To combine and refine the silhouettes and skeletons, we introduce two methods, \textit{GaitMix} and \textit{GaitRef}. \textit{GaitMix} takes  skeletons and silhouettes as inputs and includes the encoded embeddings of both modalities for end-to-end gait recognition.
\textit{GaitMix} has two encoders, a silhouette feature encoder and a skeleton feature encoder, to project two modalities to their corresponding embedding spaces, followed by an MLP layer for fusing the concatenated feature to the identity embedding space. In our experiment, \textit{GaitMix} works as the baseline since the existing state-of-the-art methods usually take only silhouette \cite{chao2019gaitset,fan2020gaitpart,lin2021gaitgl,hou2020gln} or skeleton \cite{liao2020model,teepe2021gaitgraph,teepe2022towards} as the input of the network.

Based on the features encoded from \textit{GaitMix}, \textit{GaitRef} further refines the positions of the joints in the input skeleton sequence.
We combine the encoded silhouette features with the encoded framewise skeleton features and the original joint positions to predict the relative changes for each point in the skeletons. 
Since the gait pattern should be consistent for the same person, features from the silhouettes and skeletons describing the same walking sequence should also be consistent, making the refinement of the skeletons with encoded silhouette features possible.
In addition, the sequence-level silhouette feature helps the frame-level skeletons for each frame understand its corresponding poses without losing identity information since the temporal feature for the person is consistent and is shared across all the frames in the same walking sequence.

With the predicted change of the points, we add them back to the original skeleton sequence and use the skeleton encoder to extract the skeleton feature. We then concatenate it with the silhouette feature to predict the identity of the sequence with the refined skeletons.
We assess our method on four public datasets, CASIA-B \cite{yu2006framework}, OUMVLP \cite{takemura2018multi}, Gait3D \cite{zheng2022gait} and GREW \cite{zhu2021gait}. We show that the refined skeletons with silhouettes outperform other state-of-the-art gait recognition methods using skeletons and silhouettes as input, including \textit{BaseMix}, our baseline method.

For the refinement of the input modalities, GaitEdge \cite{liang2022gaitedge} introduced using RGB images to refine the silhouettes with the corresponding RGB images in the dataset. Due to privacy concerns, most public datasets \cite{takemura2018multi,zheng2022gait,zhu2021gait} do not provide RGB images. We only require silhouettes and skeletons that are provided by the public datasets and achieve similar or even better results. We  discuss more differences between the two methods in Sec.~\ref{sec:exp}.

In summary, our contributions are 1) we introduce \textit{GaitMix} and \textit{GaitRef}, which combine the skeletons and silhouettes as end-to-end training for the gait recognition network, 2) we apply \textit{GaitRef} for refining the skeletons with encoded silhouette features for refining skeletons without losing identity information in the sequence, and 3) we assess our model on four public datasets, CASIA-B, OUMVLP, Gait3D and GREW, and show state-of-the-art performance compared with other methods for gait recognition.

\section{Related Work}

\textbf{Gait Recognition} aims to find the corresponding identity of the person from the walking pattern. Considering the privacy issues in RGB images, gaits are usually recorded as two representations, silhouettes \cite{yu2006framework,takemura2018multi} and skeletons \cite{an2020performance}. Silhouettes record the boundary map of the human segmentation. To limit the impact of appearance variants on human shapes, researchers focus on part-based and body-shape reconstruction methods for gait recognition. GaitSet~\cite{chao2019gaitset} and GLN~\cite{hou2020gln} introduce set pooling and extract set features in the sequence. GaitPart~\cite{fan2020gaitpart} and GaitGL~\cite{lin2021gaitgl} split the image into different small patches and use local features to limit the impact of the appearance variants. In addition to directly mining identity information from silhouettes, ModelGait~\cite{li2020end}, Gait3D~\cite{zheng2022gait} and Gait-HBS \cite{zhu2023gait} focus on 3-D shape reconstruction to assist the identification from sequences. 

In addition to the mining identity from silhouette sequences, some researchers \cite{liao2020model,teepe2021gaitgraph} focus on using skeletons instead of silhouettes for gait recognition. Compared with the body contours of the silhouettes, skeletons only include the joints and can remove the impact of body shapes as well as the appearance of the person. GaitGraph~\cite{teepe2021gaitgraph} uses the HRNet~\cite{wang2020deep} for joint detections and uses the generated pose sequence for recognition. PoseGait~\cite{liao2020model} splits the gait sequence into pose, limb, angle, and motion, followed by analyzing the movements for each skeleton for these four features independently before combining them together for gait recognition. For the combination of silhouettes and skeletons, Wang \etal \cite{wang2022two} directly concatenates the two features, which still suffers from erroneous joint detections.

\textbf{Pose Estimation and Refinement} focus on extracting the human body poses and refinement. With the development of transformers, pose estimation is also transforming from CNN-based networks \cite{xiao2018simple,cao2019openpose,zhu2022temporal} to transformer backbone networks \cite{Li_2021_CVPR,li2021tokenpose,yang2021transpose,xu2022vitpose}.
Pose estimation has experienced rapid development from CNNs~\cite{xiao2018simple} to vision transformer networks. Early works treat the transformer as a better decoder~\cite{Li_2021_CVPR,li2021tokenpose,yang2021transpose,YuanFHLZCW21}. Although the frame-level pose estimation accuracy is becoming more and more accurate, directly applying these methods to tasks with solid temporal relations, such as gait recognition, may introduce extra uncertainty with inaccurate joint predictions. For the sequence with strong temporal patterns, HuMoR \cite{rempe2021humor} corrects the joint prediction of the person with the previous pose, and SmoothNet~\cite{zeng2021smoothnet} filters the jitters in the whole sequence with analysis for the first and second deviation of the position for each point. These methods can fix some slight jitters in the long sequence but still suffer when the poses for a long sequence are inaccurate. For the task of gait recognition with temporal repeated patterns, even with inaccurate predictions for the long sequence, the model should still fix the joints with the consistent moving pattern of the same person, which these existing methods cannot achieve.

\section{Methods}

\begin{figure*}[t]
    \centering
    \includegraphics[width=.88\linewidth]{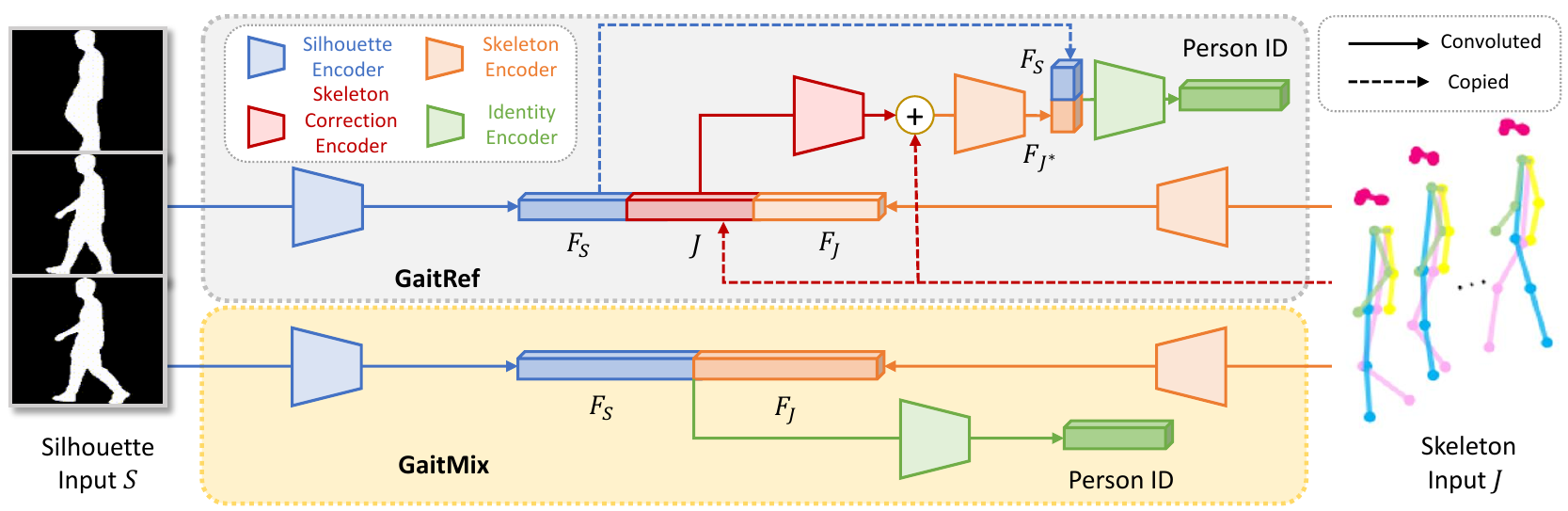}
    \caption{Our proposed architecture for GaitRef and GaitMix. Trapezoids are trainable modules, and modules of the same color in the same model share the weight. Dashed lines are the operation of feature copying. $S$ and $J$ are the input silhouettes and skeletons. $F_S$ represents silhouette features, while $F_J$ and $F_{J^*}$ represent skeleton features from input and refined skeletons, respectively. }
    \label{fig:pipeline}
\end{figure*}

Given silhouettes $S$ and joints $J$ for the person $p$, the task of gait recognition is to match the identity with the people in a pool $P=\{p_n\}_{n=1,2,...}$, where $n$ is the candidate identity. We encode $S$ and $J$ to their corresponding embeddings and find the nearest sample in $P$ in the embedding space. In this section, we discuss the details of our proposed baseline \textit{GaitMix}, which combines these two modalities, and the proposed method \textit{GaitRef} to refine the input skeleton for gait recognition. We show both architectures in Figure~\ref{fig:pipeline}. 

In the remaining of this section, we first introduce \textit{GaitMix} and \textit{GaitRef} in Sec.~\ref{sec:gaitmix} and \ref{sec:gaitref}, followed by the objectives for training in Sec.~\ref{sec:obj}.

\subsection{{GaitMix}: Multimodal Gait Recognition}\label{sec:gaitmix}

\textit{GaitMix} combines the skeletons and silhouettes as an end-to-end network for gait recognition. To extract the information from both modalities, we apply two encoders: a silhouette feature encoder for encoding the silhouette $S$, and a skeleton feature encoder for projecting the input raw skeletons $J$ into the embedding space.

\textbf{Silhouette Feature Encoder.} To extract the identity features from input sequential silhouette sequences, we use a silhouette feature encoder to convert the input silhouette sequence $S$ to the corresponding output identity feature $F_S$. We have three steps for the silhouette feature encoder: convolution feature extraction, temporal pooling, and horizontal pooling. With the binary silhouette input sequence $S = \{s_i\}_{i=1,..,N}$, where $i$ is the temporal stamp and $N$ is the overall frame number, we apply a convolution network to extract the framewise feature $f_{i}$ at frame $i$. $f_i$ is an $M$-by-$ N $-by-$ C$ matrix, where $M$ and $N$ are the height and width of the convoluted output features, and $C$ is the channel number from the output of the last convolution layer.

With the framewise feature $f_i$, we use a max pooling layer for the temporal fusion and combine the feature into a single $M$-by-$ N $-by-$ C$ output as temporal pooling. Since $f_i$ still includes the spatial features for each segment, we follow \cite{chao2019gaitset,fu2019horizontal} and apply horizontal pyramid pooling with scale $S$ as 5. The output of the feature is a $2^{S-1} $-by-$ C$ feature vector after horizontal pooling. The architecture of each component can be found in the implementation details. 

\textbf{Skeleton Feature Encoder.} In addition to the silhouette encoder, we have a skeleton feature encoder run in parallel and project the input skeleton sequence $J$ to their corresponding human identification features $F_J$. For an input skeleton sequence $J=\{j_i\}_{i=1,...,N}$, where each input consists of $K$ nodes and is shown as a $K $-by-$ 2$ matrix representing the 2-D skeletons for each frame, we follow \cite{yan2018spatial} to apply spatial-temporal graph convolution network for the graphical feature extraction. By converting the input $N$-by-$ K $-by-$ 2$ to $N$-by-$ K $-by-$ C$, we average pool on the temporal and node dimensions and generate the final $C$-length vector as $F_J$ representing the feature of the sequential skeleton.

\textbf{Fusion.} With the $2^{S-1} $-by-$ C$ silhouette feature $F_S$ and $1$-by-$ C$ skeleton feature $F_J$ encoded from two different encoders, we concatenate the two features along with their first dimension and combine it to a $(2^{S-1} + 1) $-by-$ C$ vector representing the body feature. We apply a shared MLP as an identity encoder for converting each $C$-length feature into the identity feature for identification.  

\subsection{GaitRef: Refining Skeletons with Silhouettes}\label{sec:gaitref}
Instead of directly combining skeleton and silhouette for gait recognition, \textit{GaitRef} further uses the encoded feature from silhouette to improve the skeletons from the silhouette branch with temporal consistency. Since the errors in the skeleton generation are framewise jitters, temporal consistency can fix such jitters in the skeletons. In contrast, the refined skeletons can better help silhouettes ignore the appearance variants for gait recognition. Based on the architecture of \textit{GaitMix}, \textit{GaitRef} includes two external modules, a skeleton feature encoder which is shared with \textit{GaitMix} pipeline and a skeleton correction network.

\textbf{Skeleton Correction Network.} With information from the silhouette feature, we use three different features as the network's input to correct the skeleton and compute the corresponding adjustment for each point: $2^{S-1}$-by-$C$ silhouette features $F_S$, $N$-by-$K$-by-$C$ skeleton feature before pooling, and the original $N$-by-$K$-by-$2$ joint matrix $J$. $F_S$ provides the sequential information to correct the joint features $F_J$. $F_J$ provides the framewise and feature for each node to correct the corresponding position of the joint in the frame. $J$ provides the input order of the points to ensure the input and output order of the points are the same.

\begin{figure}[t]
    \centering
    \includegraphics[width=\linewidth]{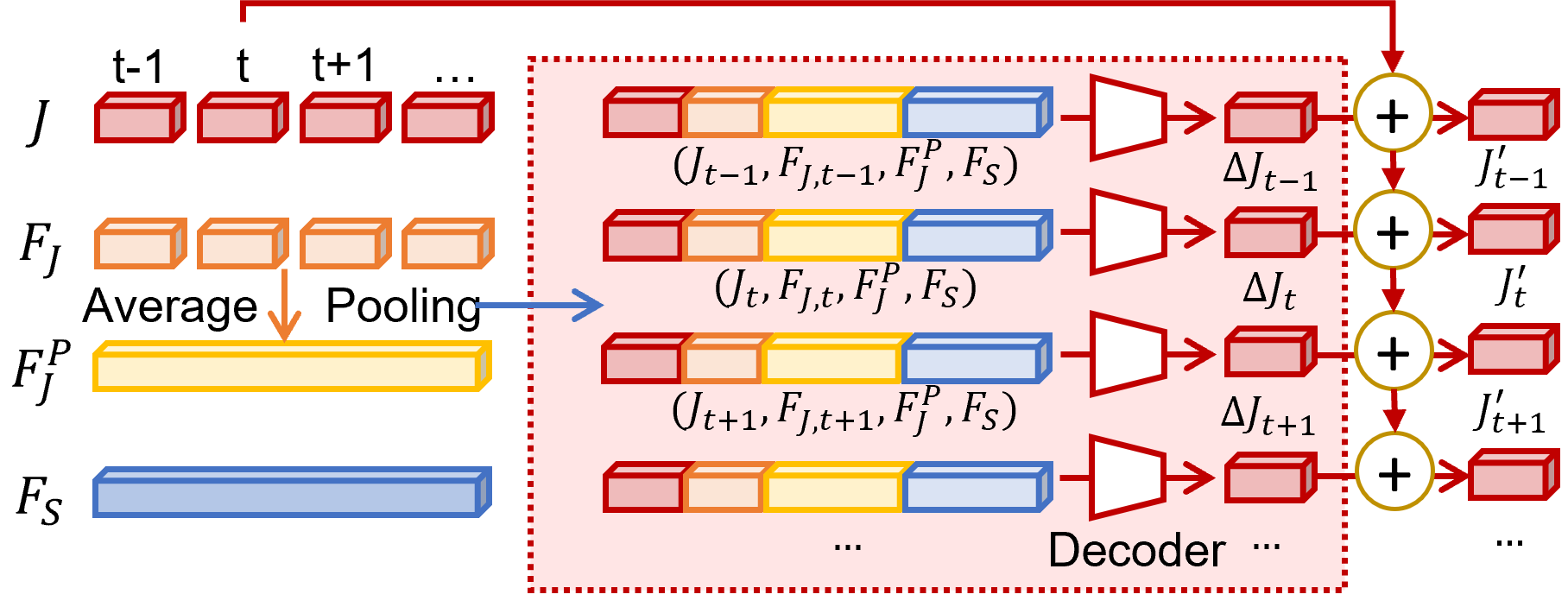}
    \caption{Architecture of the skeleton correction network. $F_J^P$ is the skeleton features after average pooling. We concatenate the joint position $J$ with its feature $F_J$ along with the global feature after pooling $F_J^P$ and the silhouette feature $F_S$ before sending it into the decoder for calculating the position difference $\Delta J$ for each frame. Decoders at different timestamps share weights.}
    \label{fig:correctnet}
\end{figure}

We show the architecture of the skeleton correction network in Figure~\ref{fig:correctnet}. With these three inputs, we first flatten the silhouette feature into a $2^{S-1}\times C$ vector. We then repeat it $N $-by-$ K$ times and concatenate it with the other two features to form a $N$-by-$K$-by-$(2^{S-1}\times C + C + 2)$ feature matrix.
To decode the new position $J'$ for each node in the sequence, we decode the $\Delta J$ for all the points with a reversed spatial-temporal graph convolution network to decode the $N$-by-$K$-by-$2$ adjustment for each node in $J$, and we have $J'$ for refine the individual points in $J$ following
\begin{equation}
    J' =J + \Delta J = J + SkeletonDecoder(J, F_S, F_J)
\end{equation}
The use of addition instead of directly predicting the corresponding location of the refined joints can give a relatively easier task for refinement and can preserve most of the original locations \cite{zhu2022open}, since the original position of the joint has most of the sequential information correct and complete. By adding $\Delta J$ on $J$, we get the final refined nodes as output and process it for further encoding.

\textbf{Skeleton Feature Encoder.} After we get the refined skeleton $J'$ with $\Delta J$, we apply the same skeleton feature encoder used for \textit{GaitMix} and apply it on the refined skeleton sequence $J'$ for predicting the $1$-by-$ C$ skeleton feature $F_{J'}$. The two skeleton feature encoders share the parameters to ensure the two embedding spaces are the same between $F_J$ and $F_J'$. Using the same skeleton feature encoder can also extend the data available for the encoder training to train a stabler graph convolution model for the feature extraction of the skeleton sequential input.

With the predicted $F_{J'}$, we concatenate it with $2^{S-1}$-by-$C$ silhouette feature $F_S$ to form a $(2^{S-1} + 1)$-by-$C$ vector for representing the human body shape for \textit{GaitRef}.

\subsection{Objectives and Inference}\label{sec:obj}
We have two losses for both \textit{GaitMix} and \textit{GaitRef}. We use a triplet loss $L_{triplet}$ for distinguishing the same identities in the same batch and a classification loss $L_{cls}$ for the identities in training set with an MLP layer for projecting the identity feature to the number of candidates. For the combination of the two losses, we follow

\begin{equation}
    L = \lambda_1 L_{triplet} +  \lambda_2 L_{cls}
\end{equation}
and empirically set $\lambda_1$ as 1. For $\lambda_2$ we follow \cite{lin2021gaitgl,zheng2022gait} to set it as different values for different datasets. We include further discussion and the choice of parameters in the implementation details section in Sec.~\ref{sec:exp}.

\section{Experiments and Results}

\subsection{Experimental Details}\label{sec:exp}

\textbf{Datasets.} In our experiment, we assess our method on four public gait recognition datasets, CASIA-B \cite{yu2006framework}, OUMVLP \cite{he2018multi,an2020performance}, Gait3D \cite{zheng2022gait} and GREW \cite{zhu2021gait}. 

\textit{CASIA-B}~\cite{yu2006framework} has 124 subjects with 10 different walking variants for gait recognition. Among the 10 variants, 6 variants are for normal walking (NM), 2 variants are for the person carrying different bags (BG), and the remaining 2 variants are for different clothes (CL). 
Each subject has 110 videos captured with 10 variants from 11 different camera viewpoints distributed between 0\degree\ and 180\degree\. 
We follow \cite{chao2019gaitset,fan2020gaitpart,hou2020gln,lin2021gaitgl} and use the videos of the first 74 identities for training and the remaining 50 for inference. 
During inference, we use the first four variances in normal walking conditions (NM) to build the gallery set as the library to query test sequences.
The sequences of the remaining 2 NM variants, along with BG and CL sequences, are used as probe examples for finding the identity in the gallery.

\textit{OUMVLP}~\cite{takemura2018multi,an2020performance} is a large-scale dataset with 10,307 different identities. Each subject in this dataset has 2 different variants for normal walking (NM) conditions from 14 camera viewpoints, making 28 gait sequences. The angles of camera viewpoints are evenly distributed in two bins, 0\degree\ to 90\degree\, and 180\degree\ and 270\degree. Every two neighbor viewpoints have a 15-degree gap. We follow \cite{chao2019gaitset,fan2020gaitpart,hou2020gln,lin2021gaitgl} to use the identities with odd indexes between the 1-\textit{st} and 10,305-\textit{th} examples and build a training set with 5,153 identities. For the remaining 5,154 identities, we use the first sequence as the gallery set and the second as probes during inference.

\textit{Gait3D} \cite{zheng2022gait} is a medium dataset compared with CASIA-B and OUMVLP for gait recognition in the wild. It includes 4,000 identities among 25,309 video sequences captured via 39 cameras. Since sequences are captured in the wild, camera positions, carried-on objects, and clothes vary from sequence to sequence. Similar to GREW \cite{zhu2021gait}, Gait3D also provides both skeletons and silhouette sequences for each frame in the dataset. We follow \cite{zheng2022gait} to use 3,000 identities for training and the remaining 1,000 during inference. For these 1,000 test cases, we build a probe set with 1,000 sequences for querying, as the probe set, and use the rest 5,369 sequences as the gallery set. 

\textit{GREW} \cite{zhu2021gait} is a large in-the-wild gait recognition dataset with 128,671 sequences capturing 26,345 identities from 882 cameras. Each frame in the video has both silhouettes and poses provided. We follow \cite{zhu2021gait} for using 20,000 identities for training and 6,000 identities as our test set. Each subject in the test set has 4 sequences, where we use two for the gallery and the other two as probes.

\textbf{Implementation Details.} For the implementation details section, we will discuss the details for the data preparation, model, and hyperparameter selection in experiments.

\textit{Data preparation.} For all four datasets, we follow OpenGait\footnote{\url{https://github.com/ShiqiYu/OpenGait}} for preparing the silhouettes for each dataset and set the size of each frame as $64\times 44$. 
Different from silhouettes, skeletons provided for different datasets are not exactly the same. Thus we process the skeletons for each dataset independently.
For CASIA-B \cite{yu2006framework} dataset, we follow GaitGraph \cite{teepe2021gaitgraph} and use a pretrained HR-Net \cite{sun2019hrnet} and generate the skeleton as MS COCO \cite{lin2014microsoft} format with 17 joints. The number of frames used for skeletons of CASIA-B is set to 60, and we use the 60 frames in the center of the whole sequence as our skeleton input. 

For OUMVLP \cite{takemura2018multi} dataset, we follow \cite{an2020performance} for applying the skeletons along with the silhouette sequences, and we have skeleton sequences with 18 nodes per frame as OpenPose \cite{cao2019openpose} format. Considering that the sequence length in OUMVLP is shorter than CASIA-B, we set the fixed frame number to 25 for each sequence. For videos shorter than 25, we repeat the frames until we have 25 frames.

For Gait3D \cite{zheng2022gait} and GREW \cite{zhu2021gait}, since skeletons are collected in the wild, we normalize each skeleton by setting their height to 2 and move their center to the origin point $(0, 0)$. This can ensure that the position of the skeletons is aligned chiefly and will not change significantly.

\begin{table*}[t]
\centering
\def\lw{0.8}
\def\la{1.5}
\def\ls{0.05}
\resizebox{0.85\linewidth}{!}
{
\begin{tabu}{p{1.5cm}p{2.7cm}p{\ls cm}p{\lw cm}<{\centering}p{\lw cm}<{\centering}p{\lw cm}<{\centering}p{\lw cm}<{\centering}p{\lw cm}<{\centering}p{\lw cm}<{\centering}p{\lw cm}<{\centering}p{\lw cm}<{\centering}p{\lw cm}<{\centering}p{\lw cm}<{\centering}p{\lw cm}<{\centering}p{\ls cm}p{\lw cm}<{\centering}} \toprule
\multirow{2}{*}{Probe}&\multirow{2}{*}{Method} && \multicolumn{11}{c}{Camera Positions} && \multirow{2}{*}{Mean} \\
 
 \cline{4-14}  \\ [-8pt]
            &&& 0\degree   & 18\degree  & 36\degree  & 54\degree  & 72\degree  & 90\degree 
            &   108\degree & 126\degree & 144\degree & 162\degree & 180\degree && \\
\midrule
\multirow{14}{*}{NM \#5-6}
         & PoseGait \cite{liao2020model}        && 55.3 & 69.6 & 73.9 & 75.0 & 68.0 & 68.2 & 71.1 & 72.9 & 76.1 & 70.4 & 55.4 && 68.7\\
         & CNN-LB \cite{wu2016comprehensive}    && 83.3 & 92.3 & 96.7 & 94.6 & 91.7 & 89.7 & 92.2 & 94.0 & 96.3 & 92.3 & 79.0 && 91.1\\
         & GaitNet \cite{song2019gaitnet}       && 91.2 & 92.0 & 90.5 & 95.6 & 86.9 & 92.6 & 93.5 & 96.0 & 90.9 & 88.8 & 89.0 && 91.6\\
         & GaitGraph \cite{teepe2021gaitgraph}  && 85.3 & 88.5 & 91.0 & 92.5 & 87.2 & 86.5 & 88.4 & 89.2 & 87.9 & 85.9 & 81.9 && 87.7\\
         & GaitSet \cite{chao2019gaitset}       && 91.1 & 98.0 & 99.6 & 97.8 & 95.4 & 93.8 & 95.7 & 97.5 & 98.1 & 97.0 & 88.2 && 95.6\\
         & GaitPart \cite{fan2020gaitpart}      && 94.0 & 98.7 & 99.3 & 98.8 & 94.8 & 92.6 & 96.4 & 98.3 & 99.0 & 97.4 & 91.2 && 96.4\\
         & GLN \cite{hou2020gln}                && 93.8 & 98.5 & 99.2 & 98.0 & 95.2 & 92.9 & 95.4 & 98.5 & 99.0 & 99.2 & 91.9 && 96.5\\
         & GaitGL \cite{lin2021gaitgl}          && 95.3 & 97.9 & 99.0 & 97.8 & 96.1 & 95.3 & 97.2 & 98.9 & 99.4 & 98.8 & 94.5 && 97.3\\
         & CSTL \cite{huang2021context}         && 97.2 & 99.0 & 99.2 & 98.1 & 96.2 & 95.5 & 97.7 & 98.7 & 99.2 & 98.9 & 96.5 && 97.8\\
         & ModelGait \cite{li2020end}           && 96.9 & 97.1 & 98.5 & 98.4 & 97.7 & 98.2 & 97.6 & 97.6 & 98.0 & 98.4 & 98.6 && \underline{\textbf{97.9}}\\
\cline{2-16}  \\ [-8pt]
         & GaitMix                              && 96.6 & 98.6 & 99.2 & 98.0 & 97.1 & 96.2 & 97.5 & 98.9 & 99.3 & 99.0 & 94.7 && 97.7 \\
         & GaitRef                              && 97.2 & 98.7 & 99.1 & 98.0 & 97.3 & 97.0 & 98.0 & 99.4 & 99.4 & 98.9 & 96.4 && \textbf{98.1} \\     
\cline{2-16}  \\ [-8pt] 
         \rowfont{\protect\leavevmode\color{gray!90}}& MvModelGait \cite{li2021end}         && 97.5 & 97.6 & 98.6 & 98.8 & 97.7 & 98.9 & 98.9 & 97.3 & 97.6 & 97.8 & 97.9 &&  {98.1}\\
         \rowfont{\protect\leavevmode\color{gray!90}}& {GaitEdge* \cite{liang2022gaitedge}}    && {97.2} & {99.1} & {99.2} & {98.3} & {97.3} & {95.5} & {97.1} & {99.4} & {99.3} & {98.5} & {96.4} && {97.9}\\
\midrule
\multirow{14}{*}{BG \#1-2}
         & PoseGait \cite{liao2020model}        && 35.3 & 47.2 & 52.4 & 46.9 & 45.5 & 43.9 & 46.1 & 48.1 & 49.4 & 43.6 & 31.1 && 44.5\\
         & CNN-LB \cite{wu2016comprehensive}    && 64.2 & 80.6 & 82.7 & 76.9 & 64.8 & 63.1 & 68.0 & 76.9 & 82.2 & 75.4 & 61.3 && 72.4\\
         & GaitNet \cite{song2019gaitnet}       && 83.0 & 87.8 & 88.3 & 93.3 & 82.6 & 74.8 & 89.5 & 91.0 & 86.1 & 81.2 & 85.6 && 85.7\\
         & GaitGraph \cite{teepe2021gaitgraph}  && 75.8 & 76.7 & 75.9 & 76.1 & 71.4 & 73.9 & 78.0 & 74.7 & 75.4 & 75.4 & 69.2 && 74.8\\
         & GaitSet \cite{chao2019gaitset}       && 87.0 & 93.8 & 94.6 & 92.9 & 88.2 & 83.0 & 86.6 & 92.6 & 95.7 & 92.9 & 83.4 && 90.1\\
         & GaitPart \cite{fan2020gaitpart}      && 89.5 & 94.5 & 95.3 & 93.5 & 88.5 & 83.9 & 89.0 & 93.6 & 96.0 & 94.1 & 85.3 && 91.2\\
         & GLN \cite{hou2020gln}                && 92.2 & 95.6 & 96.7 & 94.3 & 91.8 & 87.8 & 91.4 & 95.1 & 96.3 & 95.7 & 87.2 && 93.1\\
         & GaitGL \cite{lin2021gaitgl}          && 93.0 & 95.7 & 97.0 & 95.9 & 93.3 & 90.0 & 91.9 & 96.8 & 97.5 & 96.9 & 90.7 && \underline{94.4}\\
         & CSTL \cite{huang2021context}         && 91.7 & 96.5 & 97.0 & 95.4 & 90.9 & 88.0 & 91.5 & 95.8 & 97.0 & 95.5 & 90.3 && 93.6\\
         & ModelGait \cite{li2020end}           && 94.8 & 92.9 & 93.8 & 94.5 & 93.1 & 92.6 & 94.0 & 94.5 & 89.7 & 93.6 & 90.4 && 93.1\\
         
\cline{2-16}  \\ [-8pt]
         & GaitMix                              && 94.4 & 96.7 & 96.8 & 96.1 & 94.3 & 90.4 & 93.5 & 97.4 & 98.0 & 97.2 & 92.2 && {95.2}\\
         & GaitRef                              && 94.4 & 96.4 & 97.3 & 96.8 & 96.2 & 92.2 & 94.4 & 97.2 & 98.7 & 97.9 & 93.3 && \textbf{95.9}\\
\cline{2-16}  \\ [-8pt]
         \rowfont{\protect\leavevmode\color{gray!90}}& MvModelGait \cite{li2021end}         && 93.9 & 92.5 & 92.9 & 94.1 & 93.4 & 93.4 & 95.0 & 94.7 & 92.9 & 93.1 & 92.1 && 93.4\\
         \rowfont{\protect\leavevmode\color{gray!90}}& GaitEdge* \cite{liang2022gaitedge}    && 95.3 & 97.4 & 98.4 & 97.6 & 94.3 & 90.6 & 93.1 & 97.8 & 99.1 & 98.0 & 95.0 && {96.1}\\
\midrule
\multirow{14}{*}{CL \#1-2}
         & PoseGait \cite{liao2020model}        && 24.3 & 29.7 & 41.3 & 38.8 & 38.2 & 38.5 & 41.6 & 44.9 & 42.2 & 33.4 & 22.5 && 36.0\\
         & CNN-LB \cite{wu2016comprehensive}    && 37.7 & 57.2 & 66.6 & 61.1 & 55.2 & 54.6 & 55.2 & 59.1 & 58.9 & 48.8 & 39.4 && 54.0\\
         & GaitNet \cite{song2019gaitnet}       && 42.1 & 58.2 & 65.1 & 70.7 & 68.0 & 70.6 & 65.3 & 69.4 & 51.5 & 50.1 & 36.6 && 58.9\\
         & GaitGraph \cite{teepe2021gaitgraph}  && 69.6 & 66.1 & 68.8 & 67.2 & 64.5 & 62.0 & 69.5 & 65.6 & 65.7 & 66.1 & 64.3 && 66.3\\
         & GaitSet \cite{chao2019gaitset}       && 71.0 & 82.6 & 84.0 & 80.0 & 71.7 & 69.1 & 72.1 & 76.7 & 78.5 & 77.2 & 63.4 && 75.1\\
         & GaitPart \cite{fan2020gaitpart}      && 72.5 & 82.8 & 86.0 & 82.2 & 79.5 & 71.0 & 77.7 & 80.8 & 82.9 & 81.4 & 67.7 && 78.6\\
         & GLN \cite{hou2020gln}                && 78.5 & 90.4 & 90.3 & 85.1 & 80.2 & 75.8 & 78.1 & 81.8 & 80.9 & 83.2 & 72.6 && 81.5\\
         & GaitGL \cite{lin2021gaitgl}          && 71.7 & 90.5 & 92.4 & 89.4 & 84.9 & 78.1 & 83.1 & 87.5 & 89.1 & 83.9 & 67.4 && 83.5\\
         & CSTL \cite{huang2021context}         && 78.1 & 89.4 & 91.6 & 86.6 & 82.1 & 79.9 & 81.8 & 86.3 & 88.7 & 86.6 & 75.3 && \underline{84.2}\\
         & ModelGait \cite{li2020end}           && 78.2 & 81.0 & 82.1 & 82.8 & 80.3 & 76.9 & 75.5 & 77.4 & 72.3 & 73.5 & 74.2 && 77.6\\
         
\cline{2-16}  \\ [-8pt]
         & GaitMix                              && 79.2 & 89.5 & 94.2 & 90.0 & 84.9 & 80.3 & 85.2 & 89.2 & 90.3 & 86.9 & 73.7 && {85.8}\\
         & GaitRef                              && 81.4 & 93.3 & 94.3 & 91.6 & 87.8 & 83.9 & 88.5 & 91.7 & 91.6 & 89.1 & 75.0 && \textbf{88.0}\\
\cline{2-16}  \\ [-8pt]
         \rowfont{\protect\leavevmode\color{gray!90}}& MvModelGait \cite{li2021end}         && 77.0 & 80.0 & 83.5 & 86.1 & 84.5 & 84.9 & 80.6 & 80.4 & 77.4 & 76.6 & 76.9 && 80.7\\
         \rowfont{\protect\leavevmode\color{gray!90}}& GaitEdge* \cite{liang2022gaitedge}    && 84.3 & 92.8 & 94.3 & 92.2 & 84.6 & 83.0 & 83.0 & 87.5 & 87.4 & 85.9 & 75.0 && {86.4}\\
\bottomrule
\end{tabu}
}
\medskip
\caption{Gait recognition results on CASIA-B dataset, excluding identical-view cases. GaitEdge* requires RGB frames and uses the re-segmented CASIA-B* silhouettes instead of CASIA-B, and MvModelGait requires the input camera viewpoints. We mark the best results among all the methods in bold and the best results in our baseline methods with underline.
} 
\label{tab:casiab-1}
\end{table*}

\textit{Network details.} In our network, we have two different encoders. For our silhouette feature encoder, we follow GaitGL \cite{lin2021gaitgl} to build the encoder for CASIA-B, OUMVLP, and GREW. For Gait3D, we follow SMPLGait \cite{zheng2022gait} and use its 2-D variant baseline, which we denote as OpenGait, to encode silhouette features. 
For the silhouette feature encoder in \textit{GaitMix}, we follow ST-GCN \cite{yan2018spatial} for encoding the skeletons into the same embedding dimension $N_{out}$ as the silhouette feature encoder. The dimension of the hidden layers of ST-GCN is set to [64, 64, 128, 128, $n_{out}$]. In addition to the \textit{GaitMix}, the decoder of the \textit{GaitRef} uses the reversed shape of the ST-GCN, with [128, 64, 64, 3] as the hidden dimensions. For the encoder and decoder network, we have compared ST-GCN along with other choices, such as MS-G3D \cite{liu2020disentangling} in the ablation study.

\textit{Model training.} In our model, we follow \cite{lin2021gaitgl,zheng2022gait} for choosing the hyperparameters. For CASIA-B, OUMVLP, and GREW, we use an Adam optimizer \cite{kingma2014adam} with $1e-4$ as the learning rate for 80,000, 210,000, and 250,000 iterations, respectively. We decay the learning rate once at 70,000 iterations for CASIA-B and twice for OUMVLP and GREW, at iterations 150,000 and 200,000 as $\frac{1}{10}$ of its original value. For the Gait3D dataset, we use the Adam optimizer for 180,000 iterations and set the initial learning rate as $1e-3$, and the learning rate is decayed to $\frac{1}{10}$ three times at iteration 30,000, 90,000 and 150,000. For CASIA-B, OUMVLP and GREW, we follow \cite{lin2021gaitgl} for using 1 for $\lambda_1$ and $\lambda_2$, while we set $\lambda_2$ as 0.1 for Gait3D following \cite{zheng2022gait}.

\textit{Metrics and evaluations.} During inference, for each example in the probe set, we use $L_2$ similarity to find the nearest example in the gallery set. For CASIA-B and OUMVLP, we evaluate the top-1 accuracy for the prediction. For GREW, we evaluate top-1, 5, 10 and 20 accuracies. For Gait3D, we assess top-1 and top-5 accuracies along with mAP and mINP following \cite{ye2021deep} for assessing since all the correct matches should have low-rank values when pairing the probe example with correct identities in the gallery.

\textbf{Baseline Methods.} For baseline methods, we compare with state-of-the-art gait recognition methods, including CNN-LB \cite{wu2016comprehensive}, GaitNet \cite{song2019gaitnet}, GaitSet \cite{chao2019gaitset}, GaitPart \cite{fan2020gaitpart}, GLN \cite{hou2020gln}, GaitGL \cite{lin2021gaitgl}, ModelGait \cite{li2020end} and CSTL \cite{huang2021context}. We also compare with PoseGait \cite{liao2020model} and GaitGraph\cite{teepe2021gaitgraph}, which use skeleton sequences as the input. GaitEdge \cite{liang2022gaitedge} generates silhouettes with RGB images\footnote{\url{https://github.com/ShiqiYu/OpenGait/tree/master/datasets/CASIA-B*}} and MvModelGait \cite{li2021end} requires RGB images and camera positions, which are not provided by most of the datasets. Thus we do not make direct comparisons with them in our experiments.

\subsection{Results and Analysis}\label{sec:res}
To compare with other methods, we present both numerical results on gait recognition tasks for public datasets as well as the visualized generated skeletons from the refined branch, followed by the ablation studies.

\textbf{Numerical Results. }We show our numerical performance on the four datasets we used in Table~\ref{tab:casiab-1}, \ref{tab:oumvlp}, \ref{table:gait3d} and \ref{table:grew}. For CASIA-B and OUMVLP, identical-view cases are excluded. We have the following observations:

    \textbf{\textit{(i) Comparison with other SOTA methods.}} For all four different datasets we evaluate, we outperform the existing state-of-the-art methods with \textit{GaitRef}. In Table~\ref{tab:casiab-1}, on CASIA-B, we achieve the best performance on all splits. Specifically, on NM, BG and CL, we reduce the error rates from $2.1\%$, $5.6\%$ and $15.8\%$ to $1.9\%$, $4.1\%$, $12.0\%$, which are relatively $6.7\%$, $26.8\%$ and $25.3\%$ reduction of the error rates. Even if we compare with GaitEdge \cite{liang2022gaitedge} and MvModelGait \cite{li2021end}, which use RGB images and viewpoint angles that do not usually exist in the public dataset, \textit{GaitRef} still has a $1.6\%$ and $7.3\%$ improvement on CL, the hardest split.

    For the other three datasets, on OUMVLP in Table~\ref{tab:oumvlp}, we ties with CSTL \cite{huang2021context} for the top-1 accuracy, while we outperform it along with other methods for all the metrics on Gait3D \cite{zheng2022gait} and GREW \cite{zhu2021gait} in Table~\ref{table:gait3d} and \ref{table:grew}, which we show $2.7\%$ and $1.6\%$ improvements on Rank-1 accuracies respectively and consistent improvements on other metrics. 
    This shows the solidness of correcting the skeleton using the knowledge in the silhouette sequence as \textit{GaitRef}.
    
    \textbf{\textit{(ii) Comparison between \textit{GaitMix} and \textit{GaitRef}.}} In addition to the comparison between the existing state-of-the-art methods, we also compare the performance between \textit{GaitMix} and \textit{GaitRef}, since these two methods use the skeleton modalities and are of the same setting for a fair comparison. We note that \textit{GaitRef} outperforms \textit{GaitMix} and shows pretty consistent improvements on all splits for all four datasets. While \textit{GaitMix} introduces the skeleton information to discard the negative impact of the body contour, the inaccurate skeleton introduces some extra ambiguity, making the network unable to utilize the skeleton information maximally. With the refined skeletons, the model can capture more useful and accurate information for the identification task of the corresponding person in the video.

\begin{table*}[tb]
\centering
\def\lw{0.7}
\def\ls{0.06}
\resizebox{0.93\linewidth}{!}
{
\begin{tabular}{p{2.7cm}p{\ls cm}p{\lw cm}<{\centering}p{\lw cm}<{\centering}p{\lw cm}<{\centering}p{\lw cm}<{\centering}p{\lw cm}<{\centering}p{\lw cm}<{\centering}p{\lw cm}<{\centering}p{\lw cm}<{\centering}p{\lw cm}<{\centering}p{\lw cm}<{\centering}p{\lw cm}<{\centering}p{\lw cm}<{\centering}p{\lw cm}<{\centering}p{\lw cm}<{\centering}p{\ls cm}p{0.9 cm}<{\centering}} \toprule
\multirow{2}{*}{Method} && \multicolumn{14}{c}{Camera Positions} && \multirow{2}{*}{Mean} \\

 \cline{3-16}  \\ [-8pt]
            && 0\degree   & 15\degree  & 30\degree  & 45\degree  & 60\degree  & 75\degree  & 90\degree 
            &   180\degree & 195\degree & 210\degree & 225\degree & 240\degree & 255\degree & 270\degree && \\
\midrule
GEINet \cite{shiraga2016geinet} && 23.2 & 38.1 & 48.0 & 51.8 & 47.5 & 48.1 & 43.8 & 27.3 & 37.9 & 46.8 & 49.9 & 45.9 & 45.7 & 41.0 && 42.5 \\
GaitSet \cite{chao2019gaitset} && 79.2 & 87.7 & 89.9 & 90.1 & 87.9 & 88.6 & 87.7 & 81.7 & 86.4 & 89.0 & 89.2 & 87.2 & 87.7 & 86.2 && 87.0 \\
GaitPart \cite{fan2020gaitpart} &&  82.8 & 89.2 & 90.9 & 91.0 & 89.7 & 89.9 & 89.3 & 85.1 & 87.7 & 90.0 & 90.1 & 89.0 & 89.0 & 88.1 && 88.7 \\
GLN \cite{hou2020gln} && 83.8 & 90.0 & 91.0 & 91.2 & 90.3 & 90.0 & 89.4 & 85.3 & 89.1 & 90.5 & 90.6 & 89.6 & 89.3 & 88.5 && 89.2\\
GaitGL \cite{lin2021gaitgl} && 84.2 & 89.8 & 91.3 & 91.7 & 90.8 & 91.0 & 90.4 & 88.1 & 88.2 & 90.5 & 90.5 & 89.5 & 89.7 & 88.8 && 89.6 \\
MvModelGait \cite{li2021end} && 87.7 & 89.7 & 91.1 & 90.1 & 89.8 & 90.3 & 90.3 & 88.1 & 89.4 & 89.4 & 90.0 & 90.8 & 90.0 & 89.7 && 89.7\\
CSTL \cite{huang2021context} && 87.1 & 91.0 & 91.5 & 91.8 & 90.6 & 90.8 & 90.6 & 89.4 & 90.2 & 90.5 & 90.7 & 89.8 & 90.0 & 89.4 && \underline{\textbf{90.2}}\\
\midrule
GaitMix && 85.4 & 90.3 & 91.2 & 91.5 & 91.2 & 90.9 & 90.5 & 88.9 & 88.7 & 90.3 & 90.5 & 89.8 & 89.6 & 88.9 && {89.9}\\
GaitRef && 85.7 & 90.5 & 91.6 & 91.9 & 91.3 & 91.3 & 90.9 & 89.3 & 89.0 & 90.8 & 90.8 & 90.1 & 90.1 & 89.5 && \textbf{90.2}\\
\bottomrule
\end{tabular}
}
\medskip
\caption{Gait recognition results on OUMVLP dataset, excluding identical-view cases. } 
\label{tab:oumvlp}
\end{table*}
\begin{table}
\begin{center}
\resizebox{0.9\columnwidth}{!}
{
\begin{tabu}{p{2.4cm}p{1.2cm}<{\centering}p{1.2cm}<{\centering}p{1.2cm}<{\centering}p{1.2cm}<{\centering}}
\toprule
Methods &  Rank@1 & Rank@5 & mAP & mINP\\
 \midrule
GaitSet \cite{chao2019gaitset}&     36.70 & 58.30 & 30.01 &  17.30\\
GaitPart \cite{fan2020gaitpart} &   28.20 & 47.60 & 21.58 &  12.36\\
GLN \cite{hou2020gln} &             31.40 & 52.90 & 24.74 &  13.58\\
GaitGL \cite{lin2021gaitgl} &       29.70 & 48.50 & 22.29 &  13.26\\
OpenGait \cite{zheng2022gait} &     42.90 & 63.90 & 35.19 &  20.83\\
CSTL \cite{huang2021context} &      11.70 & 19.20 & 5.59 & 2.59 \\
SMPLGait \cite{zheng2022gait} &     \underline{46.30} & \underline{64.50} & \underline{37.16} & \underline{22.23}\\
\midrule
GaitMix & {45.80} & {65.60} & {36.74} & {22.09} \\
GaitRef & \textbf{49.00} & \textbf{69.30} & \textbf{40.69} & \textbf{25.26} \\

\bottomrule
\end{tabu}}
\end{center}
\caption{Gait recognition results reported on the Gait3D dataset with $64\times 44$ as input sizes. For all four metrics, higher values of the same metric indicate better performance.}
\label{table:gait3d}
\end{table}

\begin{table}
\begin{center}
\def\lw{1.3}
\resizebox{0.9\columnwidth}{!}
{
\begin{tabular}{p{2.5cm}p{\lw cm}<{\centering}p{\lw cm}<{\centering}p{\lw cm}<{\centering}p{\lw cm}<{\centering}}
\toprule
Methods &  Rank-1 & Rank-5 & Rank-10 & Rank-20  \\
 \midrule
PoseGait \cite{liao2020model} & 0.2 & 1.1 & 2.2 & 4.3\\
GaitGraph \cite{teepe2021gaitgraph} & 1.3 & 3.5 & 5.1 & 7.5\\
GEINet \cite{shiraga2016geinet} & 6.8 & 13.4 & 17.0 & 21.0\\
TS-CNN \cite{wu2016comprehensive} & 13.6 & 24.6 & 30.2 & 37.0\\
GaitSet \cite{chao2019gaitset}& 46.3 & 63.6 & 70.3 & 76.8\\
GaitPart \cite{fan2020gaitpart} & 44.0 & 60.7 & 67.4 & 73.5 \\
CSTL \cite{huang2021context} & 50.6 & 65.9 & 71.9 & 76.9 \\
GaitGL \cite{lin2021gaitgl} & \underline{51.4} & \underline{67.5} & \underline{72.8} & \underline{77.3} \\
\midrule
GaitMix   & {52.4} & 67.4 & {72.9} & 77.2 \\
GaitRef & \textbf{53.0} & \textbf{67.9} & \textbf{73.0} & \textbf{77.5}\\
\bottomrule
\end{tabular}}
\end{center}
\caption{Rank-1, 5, 10 and 20 accuracies on GREW dataset.}
\label{table:grew}
\end{table}

    \textbf{\textit{(iii) Comparison with using 3-D body shapes.}} Different from the other two datasets, Gait3D \cite{zheng2022gait} provides the 3-D body shapes along with silhouette sequences, which are used by SMPLGait \cite{zheng2022gait}. In Table~\ref{table:gait3d} we provide the comparison for using 3-D body shapes as SMPLGait \cite{zheng2022gait} and using skeletons as \textit{GaitMix} and \textit{GaitRef}. All these methods use OpenGait~\cite{zheng2022gait} as backbones. Compared with using silhouettes as the only input modality, the use of skeleton and body shape can both improve recognition accuracy. In SMPLGait, skeleton information is partially stored in the generated 3-D body shape for gait recognition, making \textit{GaitMix} show similar performance as SMPLGait on all four metrics.
    
Compared with SMPLGait using 3-D body shape as the second modality, \textit{GaitRef} with refined skeletons achieves a better recognition performance. Considering that the generation of the SMPL body shapes also requires skeletons \cite{sun2022putting}, inaccurate pose estimation in the 3-D body shape generation also makes the model difficult to understand the noisy body shapes with erroneous poses in SMPLGait \cite{zheng2022gait}, while \textit{GaitRef} does not suffer this with refined skeletons.

\begin{table}
\begin{center}
\resizebox{0.8\columnwidth}{!}
{
\begin{tabular}{p{1.4cm}p{2.4cm}<{\centering}p{.8cm}<{\centering}p{.8cm}<{\centering}p{.8cm}<{\centering}}
\toprule
Methods & Combination & NM & CL & BG \\
\midrule
GaitGL & N/A & 97.3 & 94.4 & 83.5 \\
\midrule
GaitMix & Padding & 96.4 & 93.7 & 83.2 \\
GaitMix & Concat. & \textbf{97.7} & \textbf{95.2} & \textbf{85.8} \\
\midrule
GaitRef & Padding & 97.5 & 94.6 & 85.8 \\
GaitRef & Concat. & \textbf{98.1} & \textbf{95.9} & \textbf{88.0} \\
\bottomrule
\end{tabular}}
\end{center}
\caption{Ablation results for different silhouette and skeleton feature combination on CASIA-B dataset for three splits. `Padding' indicates the skeleton feature is padded on each of the feature of different scales, while `concat.' means we concatenate the feature along with the scale dimension and use it only once.}
\label{table:abla-comb}
\end{table}

\textbf{Ablation studies. }For ablation studies, we present two results on 1) different ways for combining the skeleton and silhouette features, 2) different skeleton encoder and decoder networks and comparison with other skeleton refinement methods, and 3) inputs of skeleton correction network. We have both experiments conducted on the CASIA-B dataset \cite{yu2006framework} for all three different settings and present the Top-1 accuracy for the final gait recognition results. We show the results in Table~\ref{table:abla-comb}, \ref{table:abla-feat} and \ref{tab:pose} respectively and have the observations as follows:

    \textbf{\textit{(i) Feature Combination.}} In addition to concatenating the features, we also repeat and pad the skeleton feature along with each segment of the silhouette features, which we label as `padding'. We show the results in Table~\ref{table:abla-comb}. For comparison, we also add the performance of GaitGL \cite{lin2021gaitgl} in the table, which only uses the silhouette feature for gait recognition and is our backbone baseline on CASIA-B. 
  
\begin{table}
\begin{center}
\resizebox{0.94\columnwidth}{!}
{
\begin{tabular}{p{1.4cm}p{2cm}<{\centering}p{2cm}<{\centering}p{.8cm}<{\centering}p{.8cm}<{\centering}p{.8cm}<{\centering}}
\toprule
Methods & Encoder & Decoder & NM & CL & BG \\
\midrule
GaitMix & ST-GCN & N/A & 97.7 & 95.2 & 85.8  \\  %
GaitMix & MS-G3D & N/A & 98.0 & 95.5 & 86.4 \\ %
\midrule   
GaitRef & ST-GCN & ST-GCN & \textbf{98.1} & \textbf{95.9} & 88.0 \\ %
GaitRef & ST-GCN & MS-G3D & \textbf{98.1} & 95.7 & \textbf{88.5} \\ %
GaitRef & MS-G3D & ST-GCN & \textbf{98.1} & \textbf{95.9} & 88.3 \\ %
\midrule
GaitMix & \multicolumn{2}{c}{Average Smoothing} & 97.6 & 95.0 & 85.6\\
GaitMix & \multicolumn{2}{c}{Gaussian Smoothing} & 97.7 & 95.2 & 85.9 \\
GaitMix & \multicolumn{2}{c}{SmoothNet \cite{zeng2021smoothnet}} & 97.4 & 94.4 & 83.8\\
\bottomrule
\end{tabular}}
\end{center}
\caption{Ablations for different encoder and decoder combinations for \textit{GaitMix} and \textit{GaitRef} and different skeleton smoothing methods on CASIA-B datasets. Results are reported in Top-1 accuarcy.}
\label{table:abla-feat}
\end{table}

\begin{table}[t]
\centering
\def\lw{1.4}
\def\lf{1.2}
\def\ls{0.05}
\resizebox{.8\linewidth}{!}
{
\begin{tabular}{p{\lw cm}<{\centering}p{\lf cm}<{\centering}p{\lw cm}<{\centering}p{\lw cm}<{\centering}p{\lw cm}<{\centering}} 
\toprule
Split & w/o $F_J$ & w/o $F_J^P$ &  w/o $F_S$  & Full SCN \\
\midrule
NM &  97.7 & 97.9 & 97.6 & 98.1  \\
BG &  95.4 & 95.9 & 95.3 & 95.9  \\
CL &  87.0 & 87.9 & 85.9 & 88.0  \\
\bottomrule
\end{tabular}
}
\smallskip
\caption{Ablation results of different input for the skeleton correction network on CASIA-B. SCN is skeleton correction network.} 
\label{tab:pose}
\end{table}

    We note that for \textit{GaitMix}, padding the skeleton feature along with each size of the silhouette feature has worse performance compared with the GaitGL baseline even if we use an external modality, while \textit{GaitRef} has some minor improvements compared with GaitGL, indicating the raw skeleton sequence is relatively noisy and introduce external ambiguity for gait recognition if concatenated with all scales of features without refinement. Nevertheless, concatenating the skeleton sequence only once, along with the silhouette, has better performance for both \textit{GaitMix} and GaitRef. When the skeleton inputs do not dominate the feature input, the silhouette features can provide useful information from the noisy data before and after refinement.
    
    \textbf{\textit{(ii) Encoder-decoder models.}} For the choice of encoder-decoder, we choose between two of the state-of-the-art skeleton action recognition models, ST-GCN \cite{yan2018spatial} and MS-G3D \cite{liu2020disentangling}. We show the results in Table~\ref{table:abla-feat}. We show that both MS-G3D and ST-GCN can improve the performance of \textit{GaitMix} and \textit{GaitRef}. In our experiment, MS-G3D requires much larger GPU memory and at least $\times 2$ training time for each introduced MS-G3D module. Considering the similar performance and time consumption, we use ST-GCN for both encoder and decoder networks.
    
    \textbf{\textit{(iii) Different skeleton refinement methods.}} For skeleton refinement, we compare GaitRef with neighbor smoothing (average and gaussian window for neighbor three frames) and SmoothNet \cite{zeng2021smoothnet} (pretrained on H36m \cite{ionescu2013human3}) on the CASIA-B dataset with Top-1 accuracy. We show the results in Table~\ref{table:abla-feat}, where \textit{GaitRef} outperforms other methods. 
    For all three variations, 3-frame Gaussian smoothing has some small improvement compared with \textit{GaitMix} but still cannot compete with \textit{GaitRef}.
    Compared with naive temporal smoothing, which creates poses not consistent with the person in the sequence, combining the silhouette features introduces the walking patterns that do not exist in the skeletons and help them refine themselves for gait recognition. Compared with the refined skeletons from the skeleton sequence alone, external knowledge from the encoded silhouette embeddings eliminates ambiguity. It introduces ID-specific information during training when the walking pattern cannot be correctly extracted from the skeleton alone.

    \textbf{\textit{(iv) Input of skeleton correction network.}} Given the presence of three distinct inputs in our skeleton correction network besides $J$, namely $F_J$, $F_J^P$, and $F_S$, we investigate the contributions of each component and present the results in Table~\ref{tab:pose} using three splits of CASIA-B datasets. We observe that when $F_J$ and $F_S$ are excluded, there is a significant performance drop, with $F_S$ being the major contributor to the final correction. The skeleton correction network leverages temporal consistency in the skeleton sequences for correction, while the additional silhouette information offers external support for enhanced understanding. $F_P$ has limited utility, as it can be derived from $F_J$, while incorporating all three inputs leads to the best performance.

\begin{figure}[t]
    \centering
    \includegraphics[width=.9\linewidth]{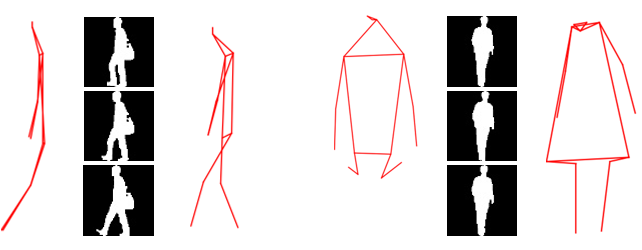}
    \caption{Visualization of successful and failure refined skeletons with \textit{GaitRef}. For each example, from left to right, we have original skeletons, silhouette of the nearby timestamp and corrected skeletons from skeleton correction network.}
    \label{fig:vis}
\end{figure}

\textbf{Skeletons visualization. } We show two examples from the \textit{GaitRef} compared to the original skeletons in Figure~\ref{fig:vis}, accompanied by the three nearest silhouettes of similar time stamp. Through the \textit{GaitRef} refinement, the resulting skeletons for gait recognition exhibit a considerable reduction in jitters and more accurately represent the individual's walking patterns, especially the visibility of feet. Although the corrected skeletons may not be entirely precise on noisier datasets (such as CASIA-B), they improve over the initial jitters by using the same skeleton encoder to align the domain between input and refined skeletons and still postively contributes to the final recognition accuracy. %

\section{Conclusion}
We introduce \textit{GaitMix} and \textit{GaitRef} for combining and refining the skeletons with silhouettes for gait recognition. \textit{GaitMix} takes skeleton and gait sequences in an end-to-end network for projecting these two modalities into the same embedding space, while \textit{GaitRef} further applies the temporal consistency in silhouettes for correcting the jitters in the skeletons. We show that combining the two modalities in \textit{GaitMix} gives more accurate predictions, while the refined skeletons with silhouettes improve the quality of skeletons and generate more precise predictions. We assess our models on four public datasets, CASIA-B, OUMVLP, Gait3D, and GREW, and show state-of-the-art performance. 

\subsubsection*{Acknowledgement}
This research is based upon work supported in part by the Office of the Director of National Intelligence (ODNI), Intelligence Advanced Research Projects Activity (IARPA), via [2022-21102100007]. The views and conclusions contained herein are those of the authors and should not be interpreted as necessarily representing the official policies, either expressed or implied, of ODNI, IARPA, or the U.S. Government. The U.S. Government is authorized to reproduce and distribute reprints for governmental purposes notwithstanding any copyright annotation therein.

{\small
\bibliographystyle{ieee}
\bibliography{egbib}
}

\end{document}